\documentclass[letterpaper]{article} % DO NOT CHANGE THIS
\usepackage{aaai2026}  % DO NOT CHANGE THIS
\usepackage{times}  % DO NOT CHANGE THIS
\usepackage{helvet}  % DO NOT CHANGE THIS
\usepackage{courier}  % DO NOT CHANGE THIS
\usepackage[hyphens]{url}  % DO NOT CHANGE THIS
\usepackage{graphicx} % DO NOT CHANGE THIS
\usepackage{natbib}  % DO NOT CHANGE THIS AND DO NOT ADD ANY OPTIONS TO IT
\usepackage{caption} % DO NOT CHANGE THIS AND DO NOT ADD ANY OPTIONS TO IT
\usepackage{amsmath}
\usepackage{amsfonts}
\usepackage{stmaryrd}
\usepackage{algorithm}
\usepackage{algorithmic}

\usepackage{booktabs}
\usepackage{amsthm}
\usepackage{titletoc}

\usepackage{newfloat}
\usepackage{listings}
\DeclareCaptionStyle{ruled}{labelfont=normalfont,labelsep=colon,strut=off} % DO NOT CHANGE THIS
\floatstyle{ruled}
\newfloat{listing}{tb}{lst}{}
\floatname{listing}{Listing}
\nocopyright
\title{Self-NPO: Data-Free Diffusion Model Enhancement via Truncated Diffusion Fine-Tuning}
\author{
Fu-Yun Wang\textsuperscript{\rm 1}, Keqiang Sun\textsuperscript{\rm 1}, Yao Teng\textsuperscript{\rm 2}, Xihui Liu\textsuperscript{\rm 2}, Jiale Yuan\textsuperscript{\rm 4}, Jiaming Song\textsuperscript{\rm 3}, Hongsheng Li\textsuperscript{\rm 1} 
}
\affiliations{
    \textsuperscript{\rm 1}CUHK, 
    \textsuperscript{\rm 2}HKU, 
    \textsuperscript{\rm 3}Luma AI, 
    \textsuperscript{\rm 4}SJTU\\
    
    fywang0126@gmail.com
}

\usepackage{bibentry}
\begin{document}

\maketitle

\begin{abstract}
Diffusion models have demonstrated remarkable success in various visual generation tasks, including image, video, and 3D content generation. Preference optimization (PO) is a prominent and growing area of research that aims to align these models with human preferences. While existing PO methods primarily concentrate on producing favorable outputs, they often overlook the significance of classifier-free guidance (CFG) in mitigating undesirable results. Diffusion-NPO addresses this gap by introducing negative preference optimization (NPO), training models to generate outputs opposite to human preferences and thereby steering them away from unfavorable outcomes through CFG. However, prior NPO approaches rely on costly and fragile procedures for obtaining explicit preference annotations (e.g., manual pairwise labeling or reward model training), limiting their practicality in domains where such data are scarce or difficult to acquire.
In this work, we propose Self-NPO, specifically \textbf{truncated diffusion fine-tuning}, a data-free approach of negative preference optimization by directly  learning from the model itself, eliminating the need for manual data labeling or reward model training. This data-free approach is highly efficient (less  than 1\% training cost of Diffusion-NPO) and achieves comparable performance to Diffusion-NPO in a data-free manner. We demonstrate that Self-NPO integrates seamlessly into widely used diffusion models, including SD1.5, SDXL, and CogVideoX, as well as models already optimized for human preferences, consistently enhancing both their generation quality and alignment with human preferences. Code is available at \url{https://github.com/G-U-N/Diffusion-NPO}.  
\end{abstract}

\begin{figure}[t]
    \centering
    \includegraphics[width=0.48\textwidth]{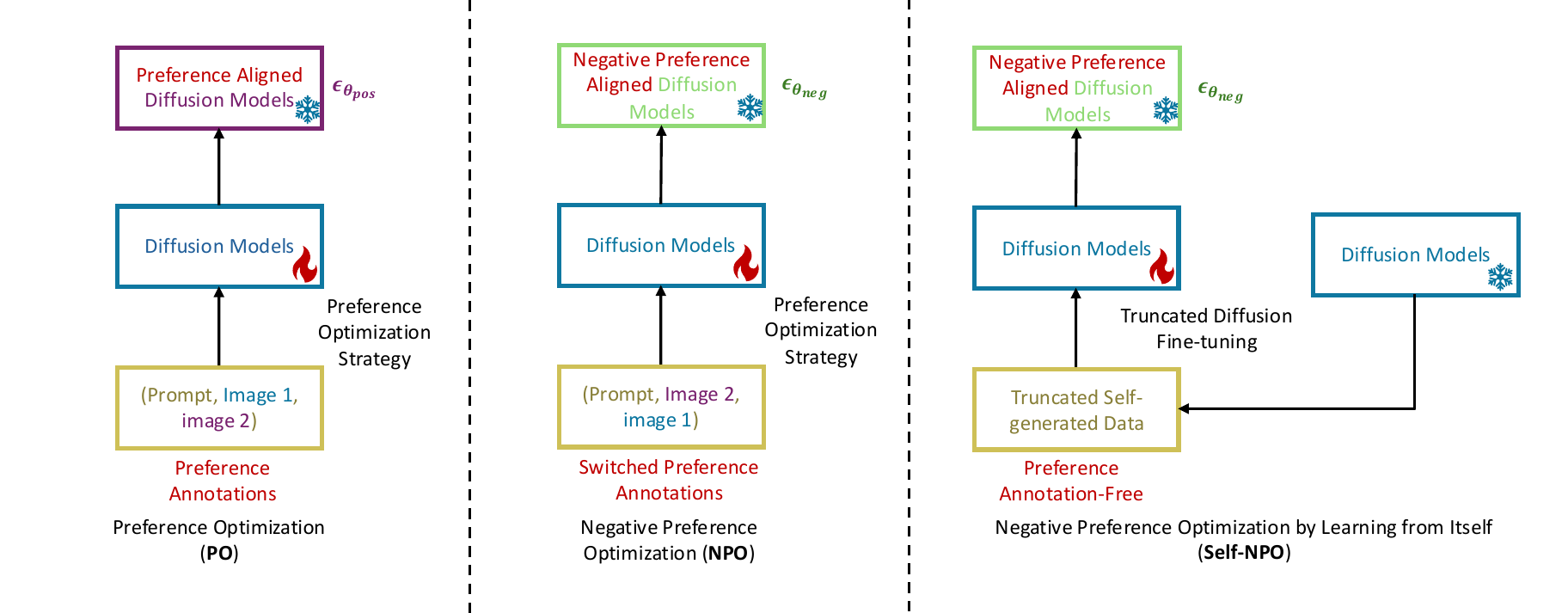}
    \caption{Motivation of Self-NPO. Preference Optimization~(PO) optimizes the preference aligned model by training on preference annotations. Negative Preference Optimization~(NPO) optimizes the negative preference aligned model by training on switched preference annotations. Self-NPO optimizes the negative preference aligned model by truncated diffusion fine-tuning~(TDFT) on truncated self-generated data.} 
    \label{fig:motivation}
\end{figure}
\section{Introduction}\label{sec:intro}
\begin{figure*}[!t]
    \centering
\includegraphics[width=0.9\linewidth]{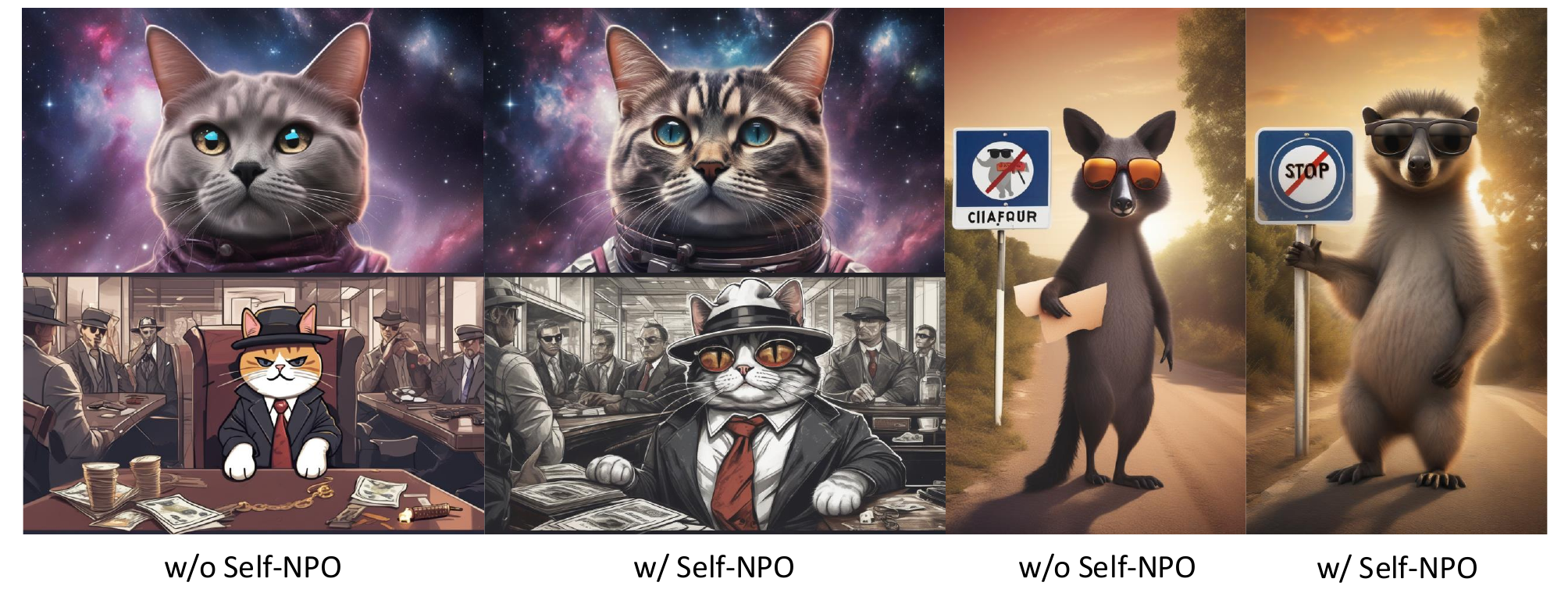}
    \caption{Visual Comparison. Prompts from  left to right, from top to bottom are ``galaxy cat'', ``cat as a mafia boss'', and ``A real picture of an Canguru with sunglasses at the sunset holding a stop sign''. }
    \label{fig:demo-2}
\end{figure*}

Over the past few years, diffusion models~\cite{ho2020denoising,song2020denoising,song2020score} have made significant strides in various visual generation tasks, including image~\cite{karras2022elucidating,rombach2022high,wang2024rectified,podell2023sdxl,diffusionbeatgan,meng2021sdedit,teng2024dim,ma2024storynizor,li2024ecnet,sun2024journeydb}, video~\cite{blattmann2023align,mao2024osv,shi2024motion,makeavideo,chen2024videocrafter2,bian2025gs,NEURIPS2022_f7f47a73,wang2024animatelcm,wang2025your,wang2024phased}, and 3D generation~\cite{gao2024cat3d,poole2022dreamfusion,li2025connecting,yan2025consistent,sun2024genca,lai2025unleashing}. However, diffusion models trained on massive unfiltered data~(\textit{e.g.}, image-text pairs) often generate results that do not align well with human preferences. The growing interest in aligning these models for human preference-aligned generation has led to the development of preference optimization~(PO) methods.  In general, current PO methods can be categorized into four types: ($\mathrm{i}$) Differentiable Reward (DR) evaluates images from iterative denoising using a pre-trained reward model, refining diffusion models through gradient-based backpropagation to align with the reward model~\cite{xu2024imagereward,prabhudesai2024video,zhang2024large,wu2023hpsv1,wu2023human,wu2024deepreward,clark2023directly}. ($\mathrm{ii}$) Reinforcement Learning (RL) models the denoising process as a Markov decision process, using techniques like PPO to optimize preferences by dynamically generating and evaluating images to maximize rewards~\cite{puterman2014markov,sutton2018reinforcement,schulman2017proximal}. ($\mathrm{iii}$) Direct Preference Optimization (DPO) streamlines training by fine-tuning with paired preference datasets, eliminating separate reward models or online evaluation, often using high-quality data as positive examples and self-generated data as negatives~\cite{rafailov2024direct,wallace2024diffusion,yuan2024self,deng2024enhancing,gu2024diffusion,zhang2024seppo}. ($\mathrm{iv}$) Negative Preference Optimization (Diffusion-NPO) trains models to avoid poor outputs by opposing human preferences, improving performance through classifier-free guidance~(CFG) to steer away from undesired results~\cite{wang2025diffusionnpo,ho2022classifier,karras2024guiding,Shen_2024_CVPR,ahn2024self}.

However, we argue that a major limitation of existing methods is their reliance on data sources, \textit{ie}, \textbf{\textit{explicit preference annotations}}, such as the expensive manual collection of preference data pairs~\cite{kirstain2023pick,wu2023hpsv2} and the challenging and fragile process of reward model training~\cite{xu2024imagereward,wu2023hpsv2,schuhmann2022laion}. This heavy dependency on preference annotations makes these approaches less practical, particularly in domains where obtaining or labeling preference data is expensive or infeasible.

In this work, we introduce \textbf{Self-NPO}, specifically, truncated diffusion fine-tuning~(TDFT), a novel approach for negative preference optimization~(NPO) that can be done in a data-free manner.   Specifically, our training objective is to apply negative preference optimization to a pre-trained model, making its predictions more aligned with the opposite direction of human preferences. This facilitates unconditional/negative-conditional outputs in classifier-free guidance, thus reducing the probability of generating results that do not align with human preferences.

To implement Self-NPO, we have three key observations:
\begin{enumerate}
    \item \textbf{Controlled weakening of generative ability.} People typically do not prefer corrupted generated images. For example, blurry details or structural errors make an image less appealing. Therefore, weakening the generative ability of a pre-trained diffusion model is, to some extent, equivalent to negative preference optimization. However, this weakening of generative ability cannot be arbitrary. In classifier-free guidance, we typically have the equation
\begin{equation}
\boldsymbol \epsilon^{\omega} =  (\omega+1)  \boldsymbol \epsilon_{\boldsymbol \theta_{pos}}(\mathbf{x}_t, t, \boldsymbol{c}) - \omega \boldsymbol \epsilon_{\boldsymbol \theta_{neg}}(\mathbf{x}_t, t, \boldsymbol{c'}) 
\end{equation}
To avoid altering the variance of the epsilon predictions, $ \boldsymbol \epsilon_{\boldsymbol \theta_{neg}}(\mathbf{x}_t, t, \boldsymbol{c'}) $ should maintain a certain level of correlation with $ \boldsymbol \epsilon_{\boldsymbol \theta_{pos}}(\mathbf{x}_t, t, \boldsymbol{c}) $. For example, for two completely independent Gaussian noises, the above operation would result in a variance of $ 2\omega^2 + 2\omega + 1 $.
    \item \textbf{Weakening via self-generated data.} This weakening, which satisfies the above conditions, can be achieved by learning from the model’s own generated data, which can be regarded as a target distribution regularized weakening. 
    \item \textbf{Efficient tuning without full diffusion.} We do not need to fully execute iterative diffusion generation process to implement self-learning. We propose a novel tuning strategy, termed as \textbf{truncated diffusion fine-tuning}, which allows us to fine tune diffusion models on the partially excuted iterative diffusion generation results.  This avoids the huge generation costs of preparing data for self-learning.  
\end{enumerate}

We demonstrate that Self-NPO can be seamlessly integrated with existing models, such as SD1.5~\cite{rombach2022high}, SDXL~\cite{podell2023sdxl}, and CogVideoX~\cite{yang2024cogvideox}, as well as models previously fine-tuned through preference optimization, consistently improving their alignment with human preferences.

\section{Preliminary}\label{sec:preliminary}

\noindent \textbf{Preliminary of CFG.} CFG has become a necessary and important technique for improving generation quality and text alignment of diffusion models. For convenience, we focus our discussion on the general formal of diffusion models, \textit{i.e.}, $\mathbf {x}_t = \alpha_t \mathbf{x}_0 + \sigma_t \boldsymbol{\epsilon}$~\cite{kingma2021variational}.  Suppose we learn a score estimator from a epsilon prediction neural network $\boldsymbol{\epsilon}_{\boldsymbol {\theta}} (\mathbf x_t, \boldsymbol c ,t)$, and we have $\nabla_{\mathbf x_t} \log \mathbb P_{\boldsymbol \theta} (\mathbf x_t | \boldsymbol c ; t) = - \frac{\boldsymbol \epsilon_{\boldsymbol \theta}(\mathbf x_t, t)}{\sigma_t}$. The sample prediction at timestep $t$ of the score estimator is formulated as
\begin{equation}\label{eq:basic-1}
\hat{\mathbf x}_0 = \frac{1}{\alpha_t} (\mathbf x_t + \sigma^2 \nabla_{\mathbf x_t} \log \mathbb P_{\boldsymbol \theta} (\mathbf x_t | \boldsymbol c ; t))\, .
\end{equation}
Applying the CFG is equivalent to add
an additional score term, that is, we replace $\nabla_{\mathbf x_t} \log \mathbb P_{\boldsymbol \theta} (\mathbf x_t | \boldsymbol c ; t) $ in Eq.~\ref{eq:basic-1} with the following term, 
\begin{equation}
\begin{split}
 \nabla_{\mathbf x_t} \log \mathbb P_{\boldsymbol \theta} (\mathbf x_t | \boldsymbol c ; t) + \nabla_{\mathbf x_t} \log \left[\frac{\mathbb P_{\boldsymbol \theta} (\mathbf x_t | \boldsymbol c ; t)}{\mathbb P_{\boldsymbol \theta} (\mathbf x_t | \boldsymbol c^\prime ; t)}\right]^\omega \, ,
\end{split}
\end{equation}
where $\omega$ is to control the strength of CFG, $\boldsymbol c$ and $\boldsymbol{c}^\prime$ are conditional and unconditional/negative-conditional inputs, respectively. It is apparent that the generation will be pushed to high probability region of $\mathbb P_{\boldsymbol \theta}(\mathbf x_t | \boldsymbol c ;t)$ and relatively low probability region of $\mathbb P_{\boldsymbol \theta}(\mathbf x_t | \boldsymbol c^{\prime} ;t)$. Write the above equation into the epsilon format, and then we have
\begin{equation}\label{eq:cfg}
   \boldsymbol \epsilon^{\omega}_{\boldsymbol{\theta}} = (\omega + 1){\boldsymbol{\epsilon}_{\boldsymbol {\theta}} (\mathbf x_t, \boldsymbol c ,t)} - \omega {\boldsymbol{\epsilon}_{\boldsymbol {\theta}} (\mathbf x_t, \boldsymbol c^{\prime} ,t)} \, .
\end{equation}

\noindent \textbf{Diffusion-NPO}~\cite{wang2025diffusionnpo}\textbf{.} Diffusion models use classifier-free guidance (CFG) to improve generation quality by combining conditional and negative-conditional outputs to favor preferred results. Traditional preference optimization aligns outputs with human preferences but overlooks avoiding undesirable outputs. Negative Preference Optimization (NPO), introduced in Diffusion-NPO, trains an additional model to counteract human preferences, reducing undesired outputs via CFG.

\textit{Training with Diffusion-NPO.} NPO adapts existing preference optimization methods without new datasets or strategies. For reward-based methods, the negative preference reward is:
\begin{equation}
R_{\text{NPO}}(\mathbf{x}, \boldsymbol{c}) = 1 - R(\mathbf{x}, \boldsymbol{c}),
\end{equation}
where \( R(\mathbf{x}, \boldsymbol{c}) \in [0, 1] \) is a reward model (\textit{e.g.}, HPSv2). For DPO-based methods, preference pairs \( r = (\mathbf{x}_0, \mathbf{x}_1, \boldsymbol{c}) \), with \(\mathbf{x}_1\) preferred, are reversed:
\begin{equation}
r_{\text{NPO}} = (\mathbf{x}_1, \mathbf{x}_0, \boldsymbol{c}).
\end{equation}

\textit{Inference with Diffusion-NPO.} 
Let \(\boldsymbol{\theta}\) denote the base model weights, with \(\boldsymbol{\eta}\) and \(\boldsymbol{\delta}\) representing the weight offsets after positive and negative preference optimization, respectively. The preference optimized model and negative preference optimized models can be represented as:
$\boldsymbol{\theta}_{pos} = \boldsymbol{\theta} + \boldsymbol{\eta}, \quad \boldsymbol{\theta}_{neg} = \boldsymbol{\theta} + \boldsymbol{\delta}.
$
Classifier-free guidance is then applied as:
$
\boldsymbol{\epsilon}^{\omega}_{\boldsymbol{\theta}} = (\omega + 1){\boldsymbol{\epsilon}_{\boldsymbol {\theta}_{pos}}} - \omega{\boldsymbol{\epsilon}_{\boldsymbol {\theta}_{neg}}}.
$
However, this often results in significant output discrepancies. To mitigate this, Diffusion-NPO modifies the negative weights to include a combination of the positive and negative offsets:
\begin{equation}
\boldsymbol{\theta}_{neg} = \boldsymbol{\theta} + \alpha \boldsymbol{\eta} + \beta \boldsymbol{\delta}, \quad \alpha, \beta \in [0, 1],
\end{equation}
which ensures more stable and correlated outputs.

\section{Methodology}\label{sec:method}

\begin{figure}[t]
    \centering
    \includegraphics[width=0.95\linewidth]{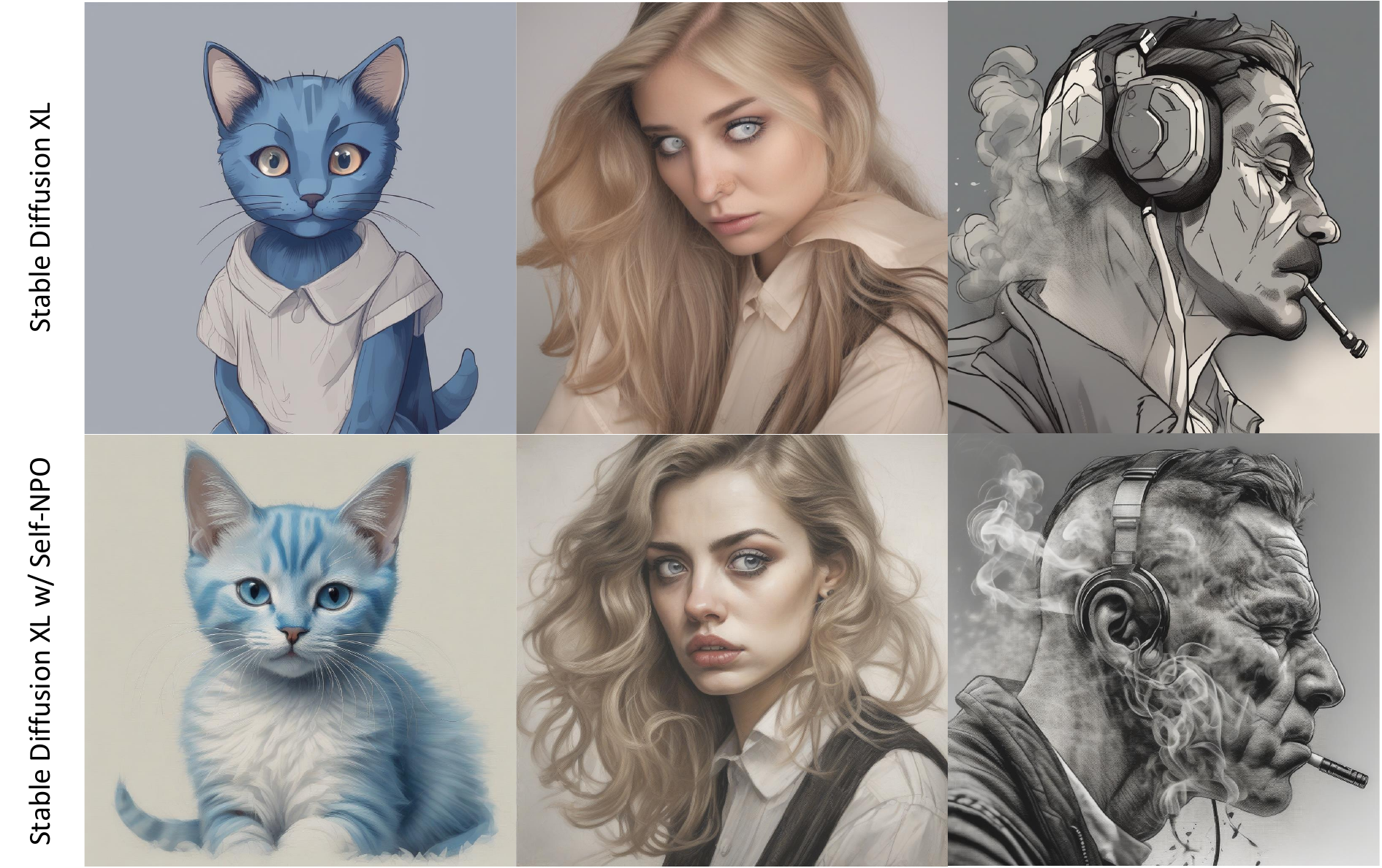}
    \caption{Visual comparison on Stable Diffusion XL. Prompts: ``A cute blue cat.'', ``An attractive young woman rolling her eyes.'', and ``Close-up of a head with smoke from ears, watching a smartphone, dynamic angle''.}
    \label{fig:demo-1}
\end{figure}

\begin{figure*}[t]
    \centering
\includegraphics[width=0.8\textwidth]{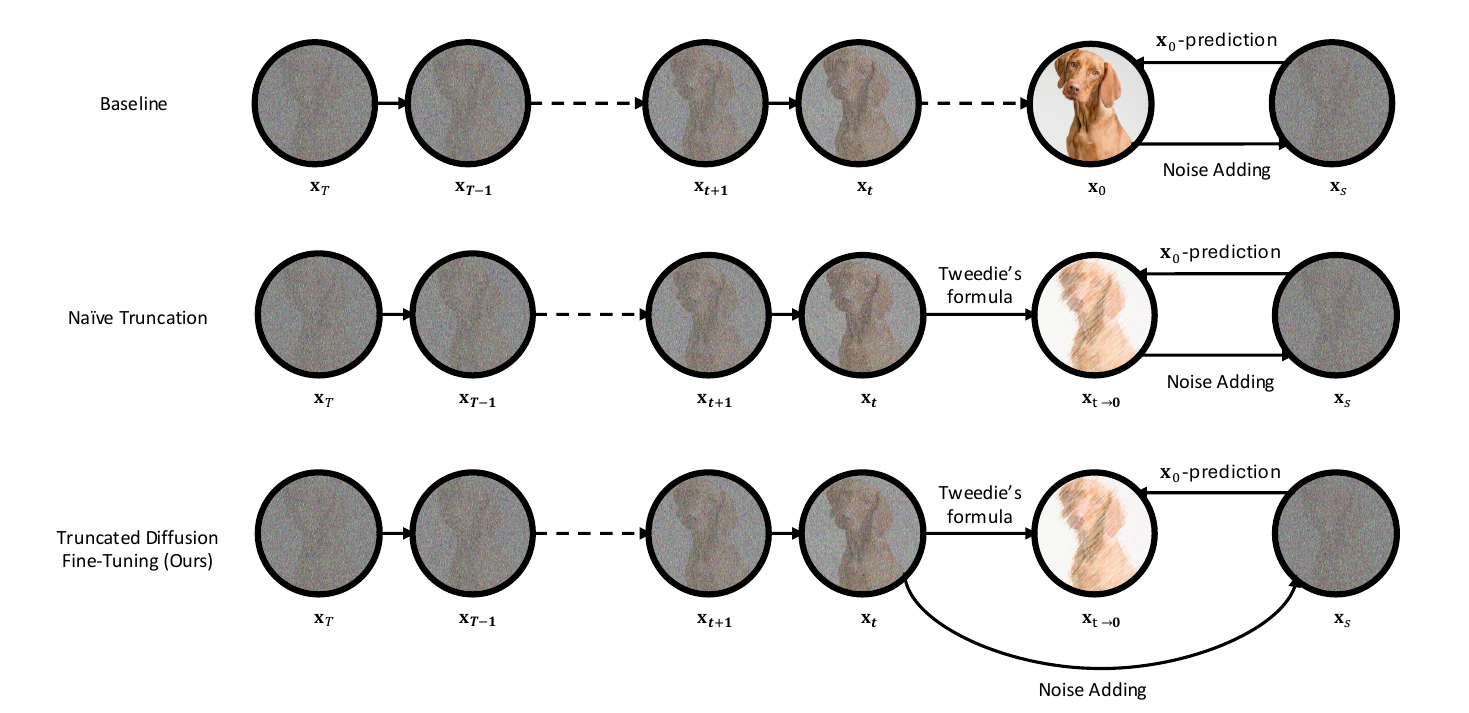}
    \caption{Truncated diffusion fine-tuning~(TDFT). Baseline method requires full simulation of generation process, introducing huge amount of generation costs. Naive truncation reduces the simulation costs but introduces distribution mismatch. TDFT reduces the simulation costs and maintains the target distribution.}
    \label{fig:method}
\end{figure*}

\subsection{NPO via self-generated data}
Reinforcement Learning with Human Feedback (RLHF)~\cite{rafailov2024direct,wallace2024diffusion,griffith2013policy,liu2025improving} aims to optimize a conditional distribution \( \mathbb{P}_\theta(\mathbf{x}_0\mid\boldsymbol{c}) \), where \( \boldsymbol{c} \sim \mathcal{D}_c \), in order to maximize the expectation value of the associated reward model \( R(\mathbf{x}_0,\boldsymbol{c}) \). Simultaneously, the optimization imposes a regularization term that penalizes deviations from a reference distribution \( \mathbb P_{\text{ref}}(\mathbf{x}_0 \mid \boldsymbol{c}) \). Formally, the objective function is formulated as follows:
\begin{equation}\label{eq:rlhf}
\begin{split}
\max_{\mathbb P_\theta} \mathbb{E}_{\boldsymbol{c} \sim \mathcal{D}_c, \mathbf{x}_0 \sim \mathbb{P}_\theta( \mathbf{x}_0 \mid \boldsymbol{c})} \left[ R( \mathbf{x}_0,\boldsymbol{c}) \right] \\ 
- \beta D_{\text{KL}} \left[ \mathbb P_\theta(\mathbf{x}_0 \mid \boldsymbol{c}) \parallel \mathbb P_{\text{ref}}(\mathbf{x}_0 \mid \boldsymbol{c}) \right] \, ,
\end{split}
\end{equation}
where \( \beta \) is a hyper-parameter that governs the balance between the reward maximization and the regularization effect imposed by the KL-divergence. The regularization is crucial because it prevents the model from straying too far from the distribution where the reward model is reliable, while also preserving generation diversity and preventing mode collapse to a single high-reward sample.
Replacing \(R(\mathbf{x}_0, \boldsymbol{c})\) with \(R_{\text{NPO}}(\mathbf{x}_0, \boldsymbol{c})\) yields the learning objective for Negative Preference Optimization (NPO):
\begin{equation}
\begin{split}
\max_{\mathbb P_\theta} \mathbb{E}_{\boldsymbol{c} \sim \mathcal{D}_c, \mathbf{x}_0 \sim \mathbb{P}_\theta( \mathbf{x}_0 \mid \boldsymbol{c})} \left[ R_{\text{NPO}}( \mathbf{x}_0, \boldsymbol{c}) \right] \\ 
- \beta D_{\text{KL}} \left[ \mathbb P_\theta(\mathbf{x}_0 \mid \boldsymbol{c}) \parallel \mathbb P_{\text{ref}}(\mathbf{x}_0 \mid \boldsymbol{c}) \right] \, .
\end{split}
\end{equation}
where the first term encourages the samples generated from $\mathbb P_{\boldsymbol \theta}$ to have low reward scores, while the second term constrains the learned distribution to remain close to \(\mathbb{P}_{\text{ref}}\).

Observing that a diffusion model can naturally produce undesirable outputs (\textit{e.g.}, disordered compositions, incoherent structures, blurry details), such low-quality samples inherently yield low reward scores. Consequently, exploiting self-generated data provides a straightforward path toward distribution-regularized negative preference optimization. Fig.~\ref{fig:motivation} summarizes the core idea and situates it among related methods. This approach is justified on two grounds:
1) \textbf{Distribution preservation.} Learning from self-generated data preserves the original distributional properties of the model.
2) \textbf{Reward reduction.} Self-generated samples exhibit low reward scores for several reasons:
a) \textit{Hallucination.}
Generation of unrealistic objects, artifacts, or contextually inconsistent outputs, which result in low reward scores due to poor quality or implausibility of the generated content.
b) \textit{Mode collapse.} 
The model may fail to generate a diverse set of outputs, leading to a lack of coverage across the entire data distribution and, consequently, lower reward scores due to insufficient variety.
c) \textit{Optimization.}
Imperfect optimization within the diffusion model itself can produce suboptimal results, such as blurry or noisy details, which leads to lower reward scores due to the misalignment between generated outputs and the desired target distribution.  \textit{Indeed, we find it is helpful to intentionally corrupt the ODE sampling process to some extent, thus increasing the probability of reward reduction. We have more discussion and empirical evidence in the supplementary material.}

However, despite its conceptual simplicity, fine-tuning a diffusion model on self-generated data can be computationally expensive, as it requires generating large amounts of training samples. To address this, we introduce a \emph{truncated diffusion fine-tuning} strategy. This method updates the model using partially generated diffusion samples, thereby avoiding the high costs of performing full diffusion simulations during training.

\subsection{Truncated diffusion fine-tuning}
\textit{Baseline: Fine-tuning on fully generated data.} We denote the reference diffusion model for data generation as $\mathbb P_{\text{ref}}(\mathbf x_0\mid \boldsymbol c)$, which models the adjacent timesteps transition conditional distribution $\mathbb P_{\text{ref}}(\mathbf x_{t-1}^{ref} \mid \mathbf x_{t}^{ref}, \boldsymbol c)$ for $t = 1, 2, \dots, T$. We can achieve data sampling by iteratively call the transition distribution, 
\begin{equation}
    \mathbf x^{ref}_{T}  \rightarrow \mathbf x^{ref}_{T-1} \rightarrow  \dots  \mathbf x^{ref}_{t} \rightarrow \dots \mathbf x^{ref}_{1} \rightarrow \mathbf x^{ref}_{0} \, ,
\end{equation}
where $\mathbf x^{ref}_{T} \sim \mathcal N (\boldsymbol 0, \mathbf I)$. After obtaining the $\mathbf x_{t}^{ref}$, the learning of $\mathbb P_{\boldsymbol \theta}(\mathbf x_0 \mid \boldsymbol c)$ is equivalent to the learning a $\mathbf x_0$-prediction neural network $\boldsymbol f_{\boldsymbol \theta}(\mathbf x_t,t, \boldsymbol c)$ with standard diffusion loss following previous work:
\begin{equation}
    \min\limits_{\boldsymbol \theta} \mathbb E_{t, \boldsymbol \epsilon} \| \mathbf x_0^{ref} - \boldsymbol f_{\boldsymbol \theta}((\mathbf x^{ref}_0)_{t}, t, \boldsymbol{c}) \|_2^2, 
\end{equation}
where $t$ is sampled from $\{1,2,\dots,T\}$, and $ (\mathbf x^{ref}_0)_{t} = \alpha_t \mathbf x^{ref}_0 + \sigma_t \boldsymbol \epsilon$.

\textit{Our method: Truncated Diffusion Fine-tuning.}
According to the Tweedie's formula, for the semi-implicit distribution \( \mathbb P_\text{ref}(\mathbf x^{ref}_t \mid \boldsymbol c)\) induced by the reference diffusion model, the expected value of \( \mathbf x_0 \) given \( \mathbf x_t^{ref} \), based on the conditional \( \mathbb P(\mathbf x_0 | \mathbf x_t^{ref}) = \frac{\mathbb P( \mathbf x_t^{ref} | \mathbf x_0) \mathbb P_{\text{ref}}(\mathbf x_0 \mid \boldsymbol c)}{\mathbb P_{\text{ref}}(\mathbf x_t^{ref} \mid \boldsymbol c)} \) according to Bayes' rule, is related to the score of \( \mathbb P_\text{ref}(\mathbf x_t^{ref} \mid \boldsymbol c) \)~\cite{efron2011tweedie,zhou2024score}. Specifically, the expectation is:
\begin{equation}
 \mathbb{E}_{\mathbf x_0 \sim \mathbb P(\mathbf x_0 | \mathbf x^{ref}_t)}[\mathbf x_0 | \mathbf x_t^{ref}] = \mathbf x_t + \sigma^2_t \nabla_{\mathbf x_t} \log \mathbb P_\text{ref}(\mathbf x_t^{ref} \mid \boldsymbol{c}). 
 \end{equation}
In this equation, the expectation \( \mathbb{E}_{\mathbf x_0 \sim \mathbb P(\mathbf x_0 | \mathbf x^{ref}_t)}[\mathbf x_0 | \mathbf x_t^{ref}] \) is computed as an integral over \( \mathbf x_0 \), and it is expressed as the current state \( \mathbf x_t^{ref} \) plus a term involving the score function \( \nabla_{\mathbf x_t^{ref}} \log \mathbb P_\text{ref}(\mathbf x_t^{ref} \mid \boldsymbol{c}) \), where \( \sigma_t^2 \) represents the noise variance. For notation simplicity, we denote $\mathbf x^{ref}_{t\rightarrow 0}$ as $\mathbb{E}_{\mathbf x_0 \sim \mathbb P(\mathbf x_0 | \mathbf x^{ref}_t)}[\mathbf x_0 | \mathbf x_t^{ref}]$ without introducing ambiguity. 

Therefore, essentially, we can obtain the expectation of $\mathbf x_0$ of $\mathbf x_t^{ref}$~(\textit{i.e.}, $\mathbf x^{ref}_{t\rightarrow 0}$) at arbitrary intermediate timesteps. Our idea is to replace the $\mathbf x_0^{ref}$ as adopted in the baseline with $\mathbf x_{t \rightarrow 0}^{ref}$, which eliminates the needs for fully simulation of generation process. Specifically, we can truncate the whole denoising process into 
\begin{equation}
        \mathbf x^{ref}_{T}  \rightarrow \mathbf x^{ref}_{T-1} \rightarrow  \dots  \mathbf x^{ref}_{t}  \xrightarrow{\text{Tweedie's formula}} \mathbf x_{t\rightarrow 0}^{ref}\, ,
\end{equation}
Further more, by incorporating real data for consideration, we can further truncate the left part by replacing the denoising trajectory from $\mathbf x_{T}^{ref}$ to $\mathbf x_{t'}^{ref}$ with the noise injection from real data, \textit{i.e.},
\begin{equation}
            \mathbf x^{ref}_{t'}  \rightarrow \mathbf x^{ref}_{t'-1} \rightarrow  \dots  \mathbf x^{ref}_{t}  \xrightarrow{\text{Tweedie's formula}} \mathbf x_{t\rightarrow 0}^{ref}\, ,
\end{equation}
where $\mathbf x_{t'}^{ref}$ is obtained by adding noise from real data following the forward process $\mathbb P(\mathbf x_{t} \mid \mathbf x_0)$. When  $t' = T$ and $t = 1$,  we have $\mathbf x_{t'}^{ref} = \mathbf x_{T}^{ref} $ and $\mathbf x_{t \rightarrow 0}^{ref} = \mathbf x_{1 \rightarrow 0}^{ref} = \mathbf x_0^{ref} \ $. In this case, truncated diffusion fine-tuning essentially reduces to the baseline method. Thus, truncated diffusion fine-tuning can be viewed as a generalization of the baseline method, as it eliminates the need for fully solving the denoising process numerically, leading to significant improvements in training efficiency.

\textit{Solving the distribution discrepancy.}  It is worthy noting that  $\mathbf x^{ref}_{t \rightarrow 0}$ indeed has a different distribution with $\mathbf x_0^{ref}\sim \mathbb P_{\text{ref}}(\mathbf x_0 \mid \boldsymbol c)$. This is because $\mathbb P(\mathbf x_{t\rightarrow 0}^{ref} \mid \mathbf x_t^{ref}, \boldsymbol c)$ can be a very smooth distribution, and $\mathbf x_{t\rightarrow 0}^{ref}$ is the weighted average value of many potential $\mathbf x_0^{ref} \sim \mathbb P_{\text{ref}}(\mathbf x_0^{ref} \mid \boldsymbol c)$.   Therefore, directly adopting the vanilla diffusion fine-tuning on $\mathbf x_{t\rightarrow 0}^{ref}$ can not achieve equivalent effect as the baseline method. We solve this distribution discrepancy by adopting a different noise adding strategy. Specifically, instead of directly adding noise to $\mathbf x_{t \rightarrow 0}^{ref}$ to obtain $\mathbf x_s$, \textit{i.e.},
\begin{equation}
    \mathbf x_s = \alpha_s \mathbf x_{t\rightarrow 0}^{ref} + \sigma_s \boldsymbol \epsilon, \quad \boldsymbol \epsilon \sim \mathcal N(\boldsymbol 0, \mathbf I) \, ,
\end{equation}
we choose to obtain $\mathbf x_s$ by adding noise to $\mathbf x_{t}^{ref}$, \textit{i.e.},
\begin{equation}
        \mathbf x_s = \frac{\alpha_s}{\alpha_t} \mathbf x_{t}^{ref} + \sqrt{\sigma_s^2 - \sigma_t^2 \frac{\alpha_s^2}{\alpha_t^2}} \boldsymbol \epsilon, \quad \boldsymbol \epsilon \sim \mathcal N(\boldsymbol 0, \mathbf I) \, .
\end{equation}
Then we optimize the diffusion model $\boldsymbol f_{\boldsymbol \theta}$ (corresponding to $\mathbb P_{\boldsymbol \theta}(\mathbf x_0\mid \boldsymbol c)$) following the standard diffusion training with $\mathbf x_{t\rightarrow 0}^{ref}$ as $\mathbf x_0$-prediction target, \textit{i.e.},
\begin{equation}\label{eq:objective}
        \min\limits_{\boldsymbol \theta} \mathbb E_{\mathbf x_s \sim \mathbb P(\mathbf x_s \mid \mathbf x_t^{ref}), \, \mathbf x_{t}^{ref}\sim \mathbb P_{\text{ref}}(\mathbf x_t^{ref} \mid \boldsymbol c)} \| \mathbf x_{t \rightarrow 0}^{ref} - \boldsymbol f_{\boldsymbol \theta}(\mathbf x_s, s, \boldsymbol{c}) \|_2^2, 
\end{equation}

\subsection{Theoretical grounding}
\textit{We theoretically show the reasonability of proposed TDFT from three aspects}:
\begin{enumerate}
    \item \textbf{\textit{Distribution of $\mathbf x_s$}}. We show that the distribution of $\mathbf x_s$ is equivalent to the noise-perturbed distribution of $\mathbb P_{\text{ref}} (\mathbf x_0^{\text{ref}} \mid \boldsymbol c)$ (See Thm.~1).
    \item \textbf{\textit{Gradient equivalent optimization target}}. We show that our optimization objective, Eq.~(\ref{eq:objective}), has a gradient-equivalent learning objective, which is the expectation value of all $\mathbf x_{t\rightarrow 0}^{\text{ref}}$ (See Thm.~2).
    \item \textbf{\textit{Equivalent optimization objective of diffusion models}}. We show that the gradient-equivalent learning objective, as proved in 2), is equivalent to the learning objective of a standard diffusion model (See Thm.~3).
\end{enumerate}

\textit{\textbf{Proof Sketch}}: Due to the limited space, we encourage readers to review the detailed proofs in Thm.~1, Thm.~2, and Thm.~3. These proofs rigorously establish the equivalence of distributions, optimization objectives, and their connections to diffusion models, forming the foundation for understanding the validity and effectiveness of our proposed approach.

\section{Experiments}\label{sec:exp}

\subsection{Validation setup}\label{sec:validation-setup}

To better evaluate the performance of our method, we tested three baseline models: a) Stable Diffusion v1-5, a text-to-image generation model with approximately 860 million parameters, widely used for creating high-quality images from textual prompts in tasks like digital art and concept visualization; b) Stable Diffusion XL (SDXL), an advanced version with around 2.6 billion parameters, enabling higher-resolution image generation (\textit{e.g.}, 1024x1024) and improved compositional coherence; c) CogVideoX-5B, a video generation model with about 5 billion parameters, designed for tasks like video synthesis and frame interpolation, leveraging its capacity for temporal modeling. Therefore, our testing covers diffusion models of varying sizes and encompasses mainstream text-to-image and text-to-video tasks, ensuring a comprehensive and well-designed baseline.

\subsection{Comparison}

\begin{figure}[t]
    \centering
    \caption{Quantitative performance comparison with stable diffusion XL based models. All metrics are tested with official weights.}
    \label{tab:sdxl}
    \resizebox{0.48\textwidth}{!}{
        \begin{tabular}{l|cccc}
            \toprule
            \textbf{Method}     &  \textbf{P.S.} & \textbf{HPS.} & \textbf{I.R.} & \textbf{AES.} \\
            \hline
            SDXL & 22.06 & 28.04 & 0.6246 & 6.114 \\
            SDXL + NPO & 22.25 & \textbf{28.98} & \textbf{0.6831} & 6.136 \\
            SDXL + Self-NPO & \textbf{22.26} & 28.24 & 0.6697 & \textbf{6.226} \\
            \hline
            Diff.-DPO & 22.57 & 29.76 & 0.8624 & 6.099 \\
            Diff.-DPO + NPO & \textbf{22.69} & \textbf{30.48} & \textbf{0.9210} & 6.112 \\
            Diff.-DPO + Self-NPO & 22.67 & 29.83 & 0.8784 & \textbf{6.179} \\
            \hline
            Juggernaut & 22.66 & 30.25 & 0.9778 & 6.021 \\
            Juggernaut + Self-NPO & \textbf{22.77} & \textbf{30.56} & \textbf{0.9921} & \textbf{6.031} \\
            \bottomrule
        \end{tabular}
    }
\end{figure}

\begin{figure}[t]
    \centering
    \includegraphics[width=0.45\textwidth]{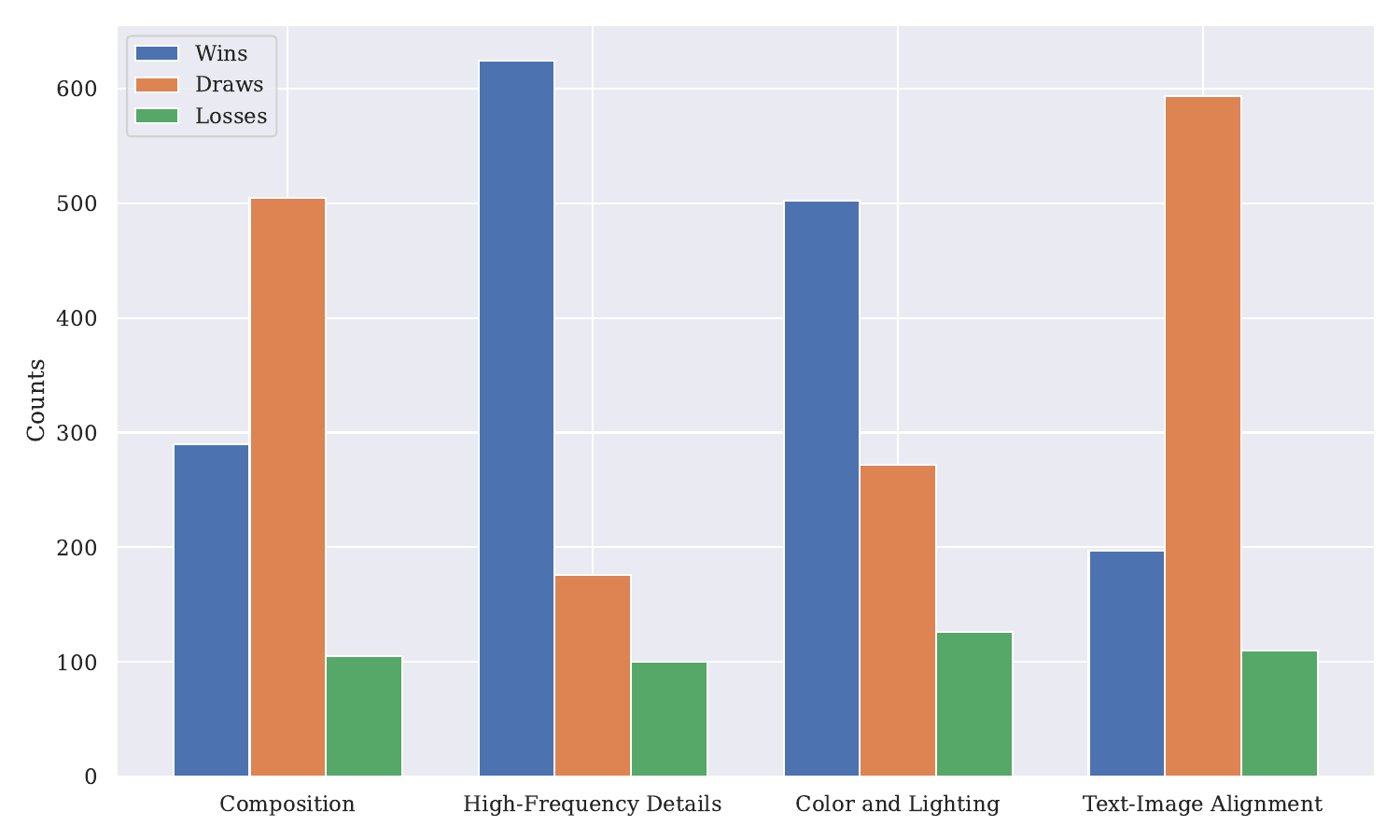}
    \caption{User study analysis.}
    \label{fig:user-study}
\end{figure}

\begin{figure*}[!t]
    \centering
    \includegraphics[width=0.95\linewidth]{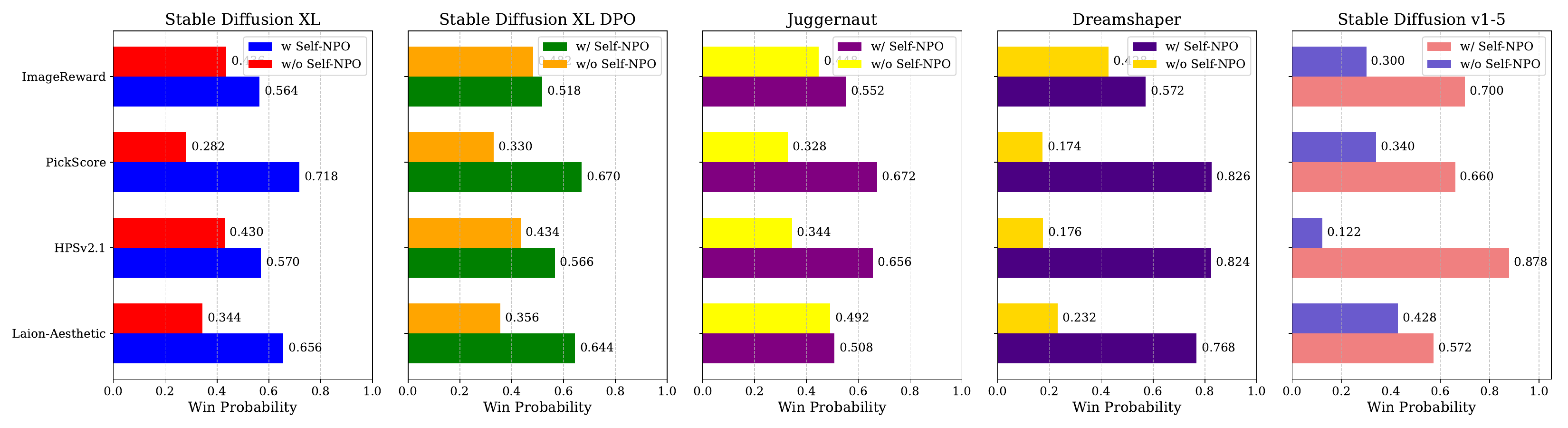}
    \caption{Winning ratio comparison. Self-NPO works seamlessly with SD1.5, SDXL, and those optimized for preferences, consistently improving both their generation quality and alignment with human preferences.}
    \label{fig:winning-ratio}
\end{figure*}
\noindent \textbf{Quantitative comparison.} For text-to-image generation, we evaluate our method quantitatively by adhering to prior research and utilizing the `test\_unique' split from Pick-a-pic as the benchmark for testing~\citep{kirstain2023pick}. We utilize PickScore~\citep{kirstain2023pick}~(\textbf{P.S.}), HPSv2.1~\citep{wu2023human}~(\textbf{HPS.}), ImageReward~\citep{xu2024imagereward}~(\textbf{I.R.}), and Laion-Aesthetic~\citep{schuhmann2022laion}~(\textbf{AES.}) as our evaluation metrics. The outcomes of this quantitative assessment are detailed in Tab.~\ref{tab:sdv15} and Tab.~\ref{tab:sdxl}. These tables reveal that integrating Self-NPO with the base model and its preference-optimized variants consistently improves the aesthetic quality of the generated images. Beyond presenting average scores, as shown in Fig.~\ref{fig:winning-ratio}, we also compute the percentage of samples generated from the same prompt that attain a higher preference score. The results produced with Self-NPO markedly surpass those generated without it. For text-to-video generation, we provide detailed experimental comparisons and demonstrations in the supplementary material, where our method achieves clear performance improvements on the majority of metrics.

\noindent \textbf{Qualitative comparison.} We present the generated results in different scenarios in Fig.~\ref{fig:demo-2}, Fig.~\ref{fig:demo-1},  and Fig.~\ref{fig:demo-application}, demonstrating that self-NPO consistently improves both the generation visual quality and fidelity.

\noindent \textbf{User study.} We evaluate generation quality across four areas: color and lighting, high-frequency details, low-frequency composition, and text-image alignment. In Color and Lighting, users assess whether the images feature natural, visually pleasing color schemes and lighting. For high-frequency details, users look at the sharpness and texture details, especially edges and fine elements. In low-frequency composition, they focus on the overall structure and balance of the image. Finally, for text-image alignment, users judge how well the generated image matches the input text. The user study uses prompts from Pickapic validation\_unique dataset, with different models generating images based on the same random seed. Users choose between ``No Preference''~(Draws), ``Self-NPO is better''~(Wins), or ``Self-NPO is worse''~(Losses) for each pair of images. We collect responses from 9 volunteers, each evaluating 100 pairs of images generated by Dreamshaper, totaling 900 votes. Results, shown in Fig.~\ref{fig:user-study}, reveal that NPO significantly improves high-frequency details, enhances color and lighting preferences, and helps with composition. It also shows an improvement in text-image alignment.

\subsection{Applications}

\begin{figure*}[t]
    \centering
    \includegraphics[width=\linewidth]{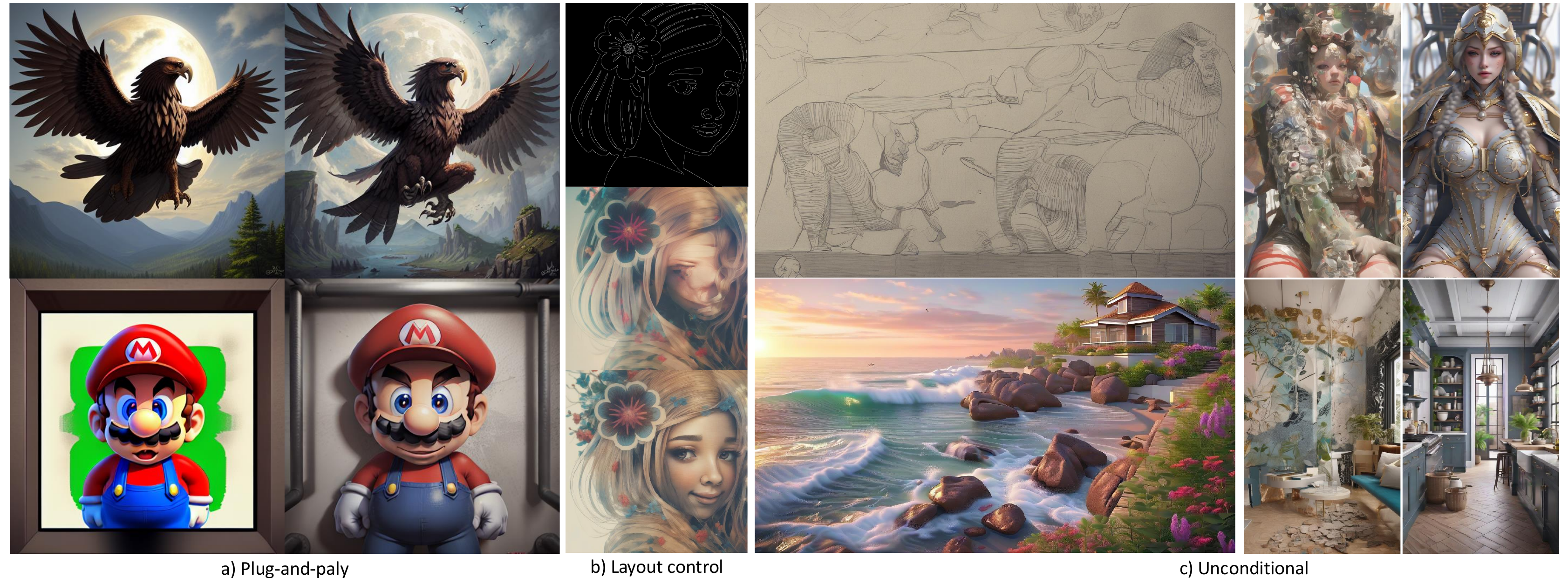}
    \caption{Applications of Self-NPO. Prompts: ``A giant eagle monster art'', ``Mario in prison'', and ``Girl, pretty, flower, hair, smile''. Results below/right use Self-NPO; above/left without it.}
    \label{fig:demo-application}
\end{figure*}

\noindent \textbf{Plug-and-play.}
Our method is not only applicable to the original stable diffusion-based models and their fine-tuned versions optimized through preference optimization, but also extends directly to high-quality stylized models fine-tuned on private datasets. As shown in Fig.~\ref{fig:demo-application}, when leveraging the Dreamshaper model alongside the self-NPO-optimized model as an unconditional predictor, we observe substantial improvements in both image quality and aesthetic appeal.

\noindent \textbf{Controllable generation.}
Moreover, our approach seamlessly integrates with controllable generation techniques. By incorporating controllable plugins like T2I-Adapter~\cite{mou2024t2i}, we significantly boost the model's generation quality under layout-controlled conditions.

\noindent \textbf{Unconditional generation.}
It is well-established that unconditional generation typically falls short of conditional generation in terms of quality. This disparity is even more pronounced in text-to-image diffusion models, where, without conditional input, the model often struggles to produce anything other than incoherent, low-quality images. However, with the self-NPO-enhanced model, we can generate relatively high-quality images even in the absence of conditional guidance.

\begin{table}[t]
    \centering
    \caption{Quantitative performance comparison with stable diffusion v1-5 based models. $^*$ means the metrics are copied from Diffusion-NPO. Other metrics are tested with official weights.}
    \label{tab:sdv15}
    \resizebox{0.99\columnwidth}{!}{  
        \begin{tabular}{l|cccc}  
            \toprule  
            \textbf{Method}     &  \textbf{P.S.} & \textbf{HPS.} & \textbf{I.R.} & \textbf{AES.} \\
            \hline  
            \textcolor{gray}{$^*$DDPO}      & \textcolor{gray}{21.06} & \textcolor{gray}{24.91} & \textcolor{gray}{0.0817} & \textcolor{gray}{5.591}\\
            \textcolor{gray}{$^*$D3PO}      & \textcolor{gray}{20.76} & \textcolor{gray}{23.97} & \textcolor{gray}{-0.1235} & \textcolor{gray}{5.527}   \\
            \textcolor{gray}{$^*$Diff.-SPO}    &\textcolor{gray}{21.41} & \textcolor{gray}{26.85} & \textcolor{gray}{0.1738} & \textcolor{gray}{5.946} \\ \hline
            SD-1.5    & 20.75 & 26.84 & 0.1064 & 5.539 \\
            SD-1.5 + NPO & 21.26 &27.36 & 0.2028 & 5.667 \\
            SD-1.5 + Self-NPO & 21.00 &27.04 & 0.2816 & 5.609 \\
            \hline
            Diff.-DPO  & 21.12 & 25.93 & 0.2651 & 5.648  \\
            Diff.-DPO + NPO~(reg$= 500$) & \textbf{21.58} & 27.60 & 0.3101 & 5.762 \\
            Diff.-DPO + NPO~(reg$= 1000$) & 21.43 & 27.36 & 0.3472 & \textbf{5.773} \\
            Diff.-DPO + Self-NPO & 21.34 & \textbf{27.78} & \textbf{0.4085} & 5.710 \\
            \hline
            SePPO & 21.51 & 28.45 & 0.5981 & 5.892  \\ 
            SePPO + Self-NPO  & \textbf{21.73}   &  \textbf{30.28}   &   \textbf{0.6744} & \textbf{6.014} \\
            \hline
            DreamShaper    &21.85 & 28.85 & 0.6819 & 6.143 \\  
            DreamShaper + NPO~($\alpha=1.0$)    &22.30 & 30.13 & 0.7258 & 6.234 \\ 
            DreamShaper + NPO~($\alpha=0.6$)    & \textbf{22.39} & 29.92 & 0.6034 & 6.492 \\ 
            DreamShaper + Self-NPO~($\alpha=1.0$)    &22.20 & \textbf{30.40} & \textbf{0.8038} & 6.196 \\ 
            DreamShaper + Self-NPO~($\alpha=0.9$)    &22.36 & 30.39 & 0.7738 & 6.345 \\ 
            DreamShaper + Self-NPO~($\alpha=0.8$)    &22.34 & 29.96 & 0.6548 & \textbf{6.562} \\
            \bottomrule
        \end{tabular}  
    }
\end{table}

\subsection{Training efficiency comparison}\label{sec:training-efficiency}
We report the training costs of Self-NPO. With the default truncated simulation steps set to 5, we trained Stable Diffusion v1-5 for 1,000 iterations with a batch size of 80. The overall training time was approximately 0.5 hours on 4 A800 GPUs (2 A100 GPU hours). For the baseline (\textit{i.e.}, full simulation, $K=25$) described in our paper, training for 1,000 iterations required approximately 2.6 hours on 4 A800 GPUs (10.4 A800 GPU hours), which is around 5 times longer than our default setting.

According to the official GitHub page of Diffusion-DPO, the official weights of Diffusion-DPO were trained with a batch size of 2,048 for 2,000 iterations, taking 24 hours on 16 A100 GPUs (384 A100 GPU hours, approximately 192 times longer than our default setting). Since DPO-based Diffusion-NPO can be trained by simply reversing the preference pair, it incurs the same training cost. However, when Diffusion-NPO is trained with LoRA, our tests show that the training time is reduced to 153.6 A800 GPU hours, approximately 76.8 times longer than our default setting.

\begin{table}[h]
\caption{Training costs for different methods.}
\label{tab:training_costs}
\centering
\resizebox{0.9\linewidth}{!}{%
\begin{tabular}{lccc}
\toprule
\textbf{Method} & \textbf{Training} & \textbf{GPU Hours} & \textbf{GPU Type} \\
\hline
Diffusion-NPO & Full weight & 384 & A100 \\
Diffusion-NPO & LoRA & 153.6 & A800 \\
Baseline & Full weight & 10.4 & A800 \\
Ours & Full weight & 2 & A800 \\
\bottomrule
\end{tabular}
}
\end{table}

\section{Conclusions}~\label{sec:conclusion}
In this paper, we explore existing diffusion model-based preference optimization techniques, including vanilla preference optimization and negative preference optimization. We observe that these methods require large amounts of explicit preference annotations or fragile reward model training, which involve costly manual data labeling and reward model training. To address this, we introduce Self-NPO (Negative Preference Optimization), a method that learns directly from data generated by the diffusion model itself, thus eliminating the need for explicit preference optimization annotations. Additionally, we propose Truncated Diffusion Fine-tuning, which reduces the dependency on full data generation simulations, significantly improving training efficiency. Extensive experimental results validate the effectiveness of Self-NPO.

\noindent \textbf{Limitations:} Since Self-NPO does not rely on explicit preference annotations, it might have a lower performance bound compared to NPO. However, it is important to highlight that by eliminating the need for explicit annotations, Self-NPO broadens the scope of preference optimization, particularly in domains where acquiring such annotations is challenging or impractical.

\paragraph{Acknowledgement.}
 This study was supported in part by National Key R\&D Program of China Project 2022ZD0161100, in part by the Centre for Perceptual and Interactive Intelligence, a CUHK-led InnoCentre under the InnoHK initiative of the Innovation and Technology Commission of the Hong Kong Special Administrative Region Government, in part by NSFC-RGC Project N\_CUHK498/24, and in part by Guangdong Basic and Applied Basic Research Foundation (No. 2023B1515130008, XW).

\bibliography{aaai2026}

\end{document}